# A real-time and unsupervised face Re-Identification system for Human-Robot Interaction

Yujiang Wang[*], Jie Shen, Stavros Petridis, Maja Pantic

*Intelligent Behaviour Understanding Group, Department of Computing, Imperial College London, London, UK*

ABSTRACT

In the context of Human-Robot Interaction (HRI), face Re-Identification (face Re-ID) aims to verify if certain detected faces have already been observed by robots. The ability of distinguishing between different users is crucial in social robots as it will enable the robot to tailor the interaction strategy toward the users' individual preferences. So far face recognition research has achieved great success, however little attention has been paid to the realistic applications of Face Re-ID in social robots. In this paper, we present an effective and unsupervised face Re-ID system which simultaneously re-identifies multiple faces for HRI. This Re-ID system employs Deep Convolutional Neural Networks to extract features, and an online clustering algorithm to determine the face's ID. Its performance is evaluated on two datasets: the TERESA video dataset collected by the TERESA robot, and the YouTube Face Dataset (YTF Dataset). We demonstrate that the optimised combination of techniques achieves an overall 93.55% accuracy on TERESA dataset and an overall 90.41% accuracy on YTF dataset. We have implemented the proposed method into a software module in the HCI^2 Framework [1] for it to be further integrated into the TERESA robot [2], and has achieved real-time performance at 10~26 Frames per second.

## 1. Introduction

Face recognition problem is one of the oldest topics in Computer Vision [3]. Recently, the interest in this problem has been revamped, mostly due to the observation that standard face recognition approaches do not perform well in real-time scenarios where faces can be rotated, occluded, and under unconstrained illumination. Face recognition tasks are generally classified into two categories:

1. Face Verification. Given two face images, the task of face verification is to determine if these two faces belong to the same person.

2. Face Identification. This refers to the process of finding the identity of an unknown face image given a database of known faces.

However, there are certain situations where a third type of face recognition is needed: face re-identification (face Re-ID). In the context of Human-Robot Interaction (HRI), the goal of face Re-ID is to determine if certain faces have been seen by the robot before, and if so, to determine their identity.

Generally, a real-time and unsupervised face re-identification system is required to achieve effective interactions between humans and robots. In the realistic scenarios of HRI, the face re-identification task is confronted with the following challenges:

a. The system needs to be able to build and update the run-time user gallery on the fly as there is usually no prior knowledge about the interaction targets in advance.

b. The system should achieve high processing speed in order for the robot to maintain real-time interaction with the users.

c. The method should be robust against high intra-class variance caused by illumination changes, partial occlusion, pose variation, and/or the display of facial expressions.

d. The system should achieve high recognition accuracy on low-quality images resulted from motion blur (when the robot and / or the user is moving), out-of-focus blur, and/or over /under-exposure.

Recently, deep-learning approaches, especially Convolutional Neural Networks (CNNs), have achieved great success in solving face recognition problems [4]–[8]. Comparing to classic approaches, deep-learning-based methods are characterised by their powerful feature extraction abilities. However, as existing works mostly focused on traditional face identification problems, the potential applications of deep-learning-based methods in solving face Re-ID problems is yet to be explored.

In this paper, we present a real-time unsupervised face re-identification system that can work effectively in an unconstrained environment. Firstly, we employ a pre-trained CNN [7] as the feature extractor and try to improve its performance and processing speed in HRI context by utilising a variety of pre-processing techniques. In the Re-Identification step, we then use an online clustering algorithm to build and update a run-time face gallery and to output the probe faces' ID. Experiments show that our system can achieve a Re-ID accuracy of 93.55% and 90.41% on the TERESA video dataset and the YTF Dataset respectively and is able to achieve a real-time processing speed of 10~26 FPS.

## 2. Related Works

Various methods [9]–[15] have been developed to solve the person Re-ID problem in surveillance context. However, most of them [9]–[13] are unsuitable to HRI applications as these approaches often rely on soft biometrics (i.e. clothing's colours and textures) that are unavailable to the robot (which usually only sees the user's face). Due to the unavailability of such soft biometrics, it is difficult to apply person re-identification

[*] Corresponding author. E-mail: yujiang.wang14@imperial.ac.uk
Code is available at: https://github.com/ibug-group/face_reid

methods in HRI scenarios, and the idea of face re-identification emerged as a result.

The idea of re-identifying, or 'remembering' different persons by their faces for humanoid robots can be traced back to the work of Aryananda [16]. That paper argued that humanoid social robots should be able to remember the identities of individual users and to learn about the different characteristics of these people. It is also pointed out that we should enable robots to actively collect training data from the environment instead of preliminarily labelling and encoding such data into the database of those robots. As a result, an online and unsupervised face recognition system for the Kismet robot [17] was proposed, where the database is initially empty and the robot collected and labelled the faces of users while interacting with them. Although the idea and target of that work is similar to ours, it requires a highly-constrained environment since it is based on the eigenface method [18], and therefore it is not suitable for applications in the wild. Also, one needs to keep interacting with robots for the face recognition performance to gradually improve, while in our work, captured faces are re-identified at real-time speed with a stable and reliable performance.

Mou's work [19] also emphasised the importance of unsupervised and online face recognition, and it demonstrated an autonomous and self-learning face recognition system regulated by a state machine. In that work, the database was also empty in the beginning and it was updated by incoming faces in an unsupervised method. A feature extraction approach based on local abstract characteristics such as lines and edges was used to obtain features of different blocks of a face image, followed by an encoding method to transform such features into a vector. In the classification step, a so-called fusing clustering algorithm, which was a combination of the classic hierarchical and partitioning algorithms, was proposed and adapted. This work extended the idea of [16] and developed a completed framework for unsupervised face recognition, however its main limitation is that it cannot satisfy the real-time requirement of HRI, as the processing speed of the whole system is 1~2 fps. A systematic evaluation on the recognition performance is also missing.

For other related works, [14] proposed a Re-ID system based on face image alone using Local Ternary Patterns (LTP) for feature extraction. However, its limitation is that the method only gives a binary output about whether a person has been seen before without further distinguishing between different users. Another face Re-ID method is presented in [15]. It specifically attempts to solve the frontal-to-side face recognition problem by combining facial features with other soft biometric information such as the colour and textural features of hair, skin and clothes. However, the method does not generalise well to HRI scenarios in which the users' head pose is not constrained to full-frontal and full-profile views.

### 3. Problem Definition

Generally Re-ID tasks could fall into two categories [9]: the close set Re-ID problem and the open set Re-ID problem. In the close set Re-ID problem, the probe is a subset of the gallery set, while for the open set Re-ID problem, the gallery set will not essentially include all the probes. Fig. 1 gives an overview of the general process of a typical open set Re-ID problem.

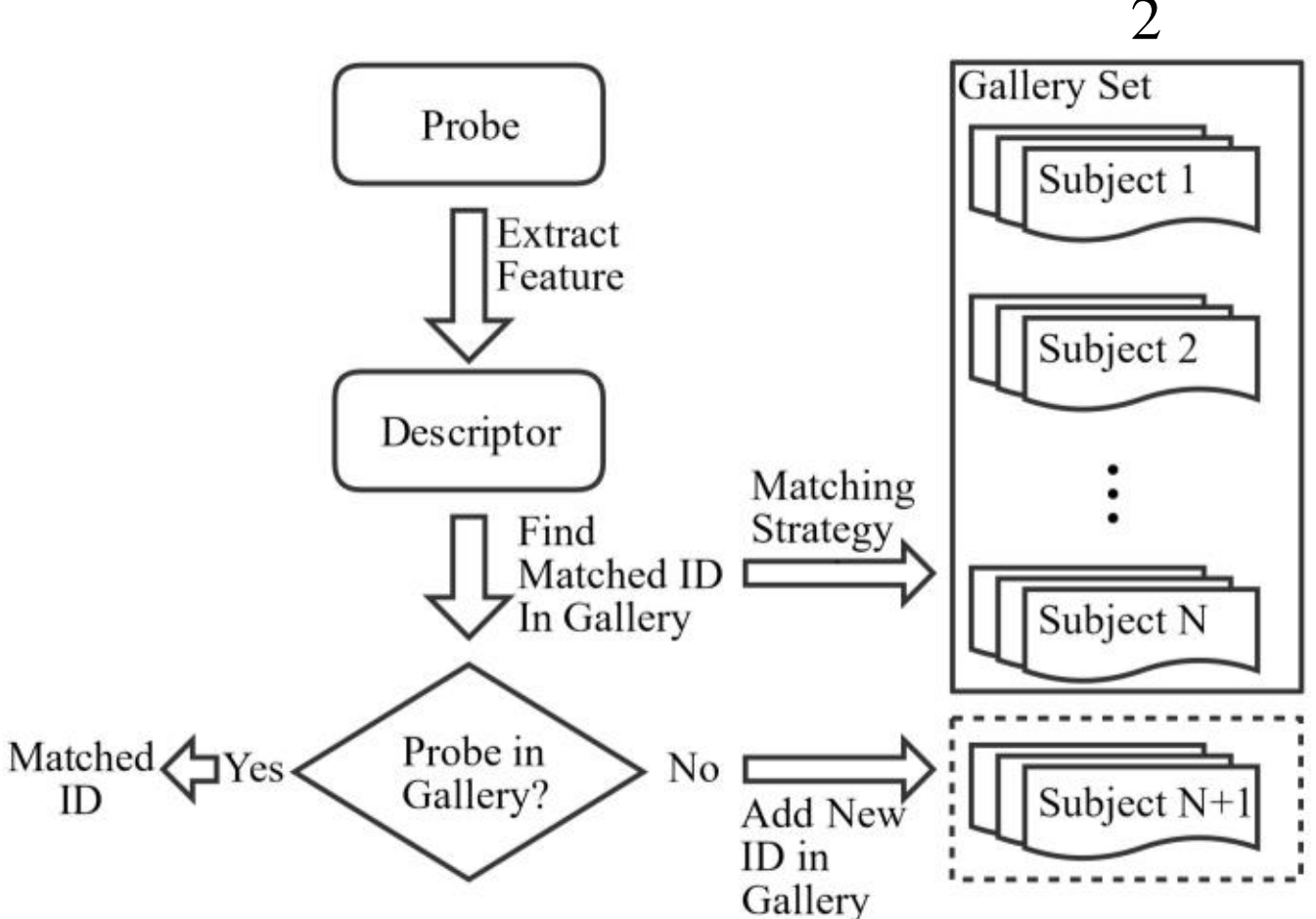

**Fig. 1.** The general process of an open set Re-ID problem

In our scenario, the users' IDs are not available in advance, which indicates that no prior gallery set can be gathered for the Re-ID system. Therefore, we deal with an open set Re-ID problem where the gallery set will dynamically increase with time. Additionally, our scenario is also a multiple Re-ID problem, since the simultaneous presence of multiple users is expected.

In most scenarios in HRI, the robot has to be close to its users. In contrast to the classic person Re-ID problems, it is difficult to obtain the global appearance of people while the users' faces are commonly accessible.

Therefore, this proposed Re-ID problem could be described as twofold: 1. Novelty Detection, 2. Re-Identification.

The novelty detection is to determine if the probe's ID could be found in the already collected gallery set. Denote the gallery set as $G = \{G_1, G_2, G_3, \ldots, G_N\}$, and the probe set $P = \{P_1, P_2, \ldots, P_M\}$. The ID for a certain subject X will be denoted as ID(X) and the IDs of the gallery set are already known, i.e. $\mathrm{ID}(G) = \{\mathrm{ID}(G_1), \mathrm{ID}(G_2), \mathrm{ID}(G_3), \ldots, \mathrm{ID}(G_N)\}$. For a probe $P_x \subseteq P$, its similarity with every subject in gallery will be measured using certain matching strategies. The gallery will be ranked according to the similarity. Specifically, in order to determine if $P_x$'s ID could be found in the gallery set, i.e. $\mathrm{ID}(P_x) \subseteq \mathrm{ID}(G)$, the novelty detection can be illustrated by Eq.(1):

$$\max_{i=1,2,\ldots,N} \left( p(\mathrm{ID}(P_x) = \mathrm{ID}(G_i)) \right) \geq \mathrm{T} \tag{1}$$

where $p(\mathrm{ID}(P_x)=\mathrm{ID}(G_i))$ refers to the likelihood that $P_x$ and $G_i$ share the same ID, and this likelihood is directly determined by their similarity, and T is a threshold value indicating if the level of similarity could be reliably trusted to assume that $\mathrm{ID}(P_x) \subseteq \mathrm{ID}(G)$.

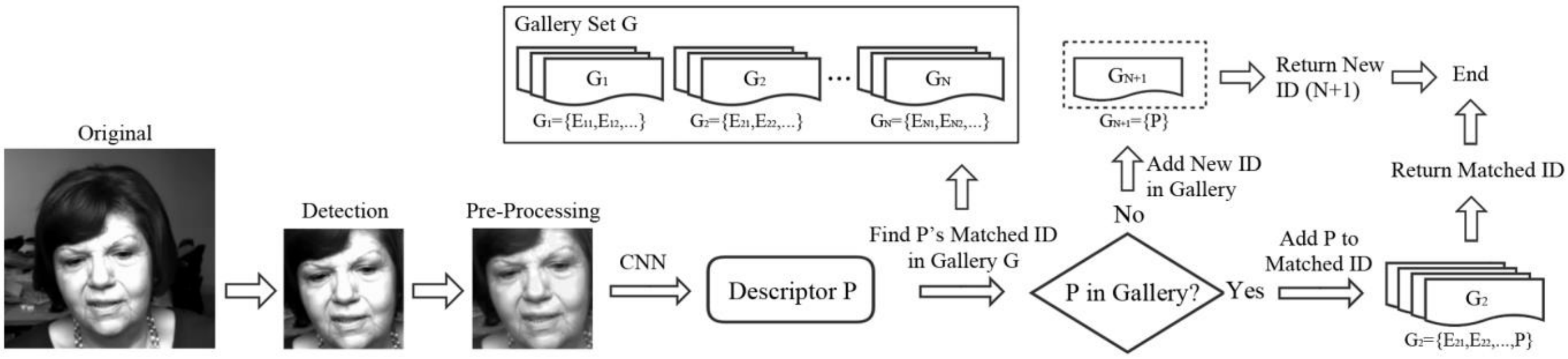

**Fig. 2.** An overview of the Re-ID system (Pre-processing includes Ghost Elimination and Histogram Equalization)

After the novelty detection, the probe's matched ID will be assigned. If Eq.(1) is not satisfied, $P_x$ will be added into the gallery set and a new ID will be registered for it. Otherwise, $P_x$'s matched ID could be obtained using Eq.(2). Eq.(2) aims to find a matched ID in the gallery set, $ID(G_{i*})$, so that the similarity between $P_x$ and $G_{i*}$ is the highest.

$$\mathrm{ID}(P_x) = \mathrm{ID}(G_{\mathrm{i}*}), \mathrm{i}* = \underset{i=1,2,\ldots,\mathrm{N}}{\arg\max}(p(\mathrm{ID}(P_x) = \mathrm{ID}(G_\mathrm{i}))) \quad (2)$$

However, we deal with a multiple Re-ID problem. Therefore, if the assigned IDs for the probe set are denoted as $ID(P) = \{ID(P_1), ID(P_2), \ldots, ID(P_M)\}$, the following condition should also be taken into consideration:

$$\underset{i=1,2,\ldots,\mathrm{M}}{\mathrm{ID}(P_i)} \neq \underset{j=1,2,\ldots,\mathrm{M}}{\mathrm{ID}(P_j)}, i \neq j \quad (3)$$

Eq.(3) is a restriction to assure that the allocated IDs for a probe set should be mutual exclusive. In practical, it guarantees that no identical ID will be assigned to two different faces simultaneously.

## 4. System Description

This section will give an overview of the Re-ID system. The general pipeline is: 1. face localisation, 2.pre-processing, 3.Re-ID.

The first step is to localise faces which are encountered by the robot. After detection, several pre-processing techniques are used to boost the Re-ID accuracy. We have also tested if the performance could be improved by applying face alignment, using the alignment techniques provided by [20]. Then the VGG-Face CNN from [7] is applied to extract the face descriptors which constitute the probe set. The next stage would be to find the probe's matched IDs in the gallery set using certain matching strategies, as mentioned in Section 3. Finally the matched IDs are returned, and the probe set will be used to update the gallery set. Fig. 2 illustrates the whole pipeline.

### *4.1. Face Localisation*

We utilise the method in [21] to do robust face localisation. First a Viola and Jones [22] face detector is used to detect the faces in front of the robot, then a face landmark tracker [21] is applied to track each face in later frames. To speed up the process, the face detector is only used when one face tracker has lost its target, i.e. the tracked face disappears.

### *4.2. Pre-Processing*

Pre-processing is necessary and essential, as our face detection system may suffer from 'ghost' detections occasionally. Fig. 3 gives an example of the 'ghost' detection: a 'ghost' face is detected where no face appears, while two other faces are located correctly. Therefore, it is crucial to eliminate such false positive faces. This is achieved by a heuristic rule, since most ghost faces last for a few frames. A detected face is considered 'ghost' and will not enter the pipeline if it has not been tracked for more than a certain time (typically one second).

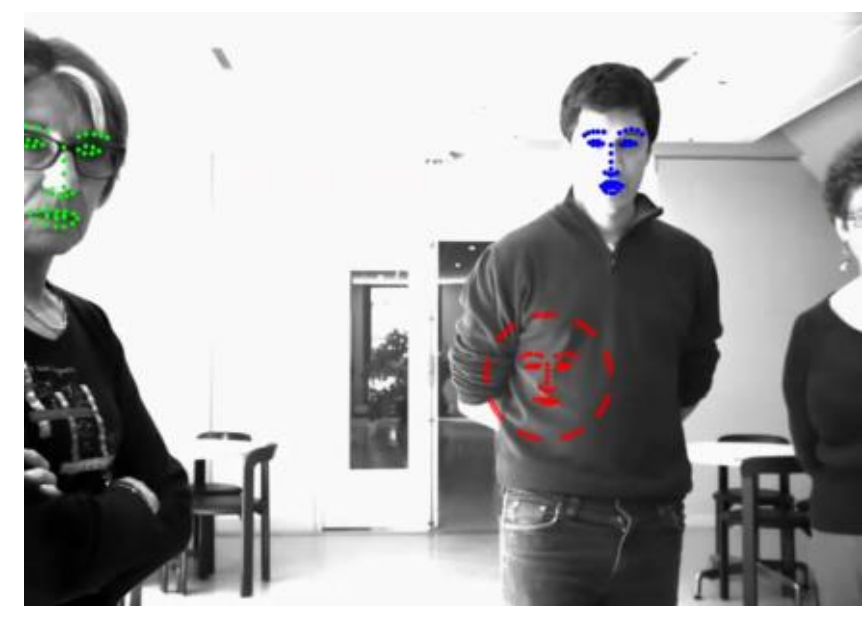

**Fig. 3.** An example of the 'ghost' detection: the face tracker in the red circle is a false positive detection (best seen in colour)

Besides, rejecting bad-quality facial images is also important since such images could lead to a gallery set containing low-quality descriptors, which will decrease the Re-ID accuracy. In this Re-ID system, the rejection of low-qualities facial images is implicitly implemented by a combination of the face tracker and ghost elimination. The face tracker typically ignores facial images whose quality is too bad to be tracked. Hence, few such images enter the pipeline, and the application of ghost elimination technique further rejects low-quality images.

We have also employed Histogram Equalization [23] which could enhance the contrast of images and eliminate the illumination variation. Besides, we have also evaluated the performance of face alignment in order to counter the effect of head pose variations, using the technique provided by [20] which uses one single 3D face surface to estimate the shape of input faces.

The last pre-processing is to resize the face images to 256 by 256, which is the required size for the VGG-Face CNN.

### *4.3. Feature Extraction*

To represent a face image with compact and discriminative descriptors has always been a core problem in face recognition. Feature extraction for faces in the wild is a challenging problem due to the high variations between intra-class instances brought by partial occlusions, facial expressions, lightings, etc. Deep-learning approaches have has great success in the past few years in extracting representative descriptors out of unconstrained faces while retaining robustness. Compared with classic handcrafted features such as Local Binary Patterns (LBP) or Scale-invariant Feature Transform (SIFT), deep-learning methods require a huge amount of data and powerful computing abilities during training.

In this Re-ID system, the VGG-Very-Deep-16 CNN architecture [7] is employed to extract the descriptor of the face images. This CNN model achieved an Equal Error Rate (EER) of 97.27% on the Labelled Face in the Wild (LFW) Dataset [24] and 92.8% EER on the YouTube Face Dataset [25] and has been trained on 2.6 million images.

The input image size of this CNN is 224 by 224. First a mean face image computed from the training set in [7] is subtracted from the input image. Then the face descriptor is obtained from the output of the penultimate layer (also the last Fully Connected Layer), which is a 4096-Dimensional Vector. Additionally, a L2-normalisation is implemented for all the descriptors in order to reduce the influence of lighting conditions.

Two different approaches are followed to extract descriptors from an image. The first approach is to crop a 224 by 224 image patch from the central area, and the central crop's descriptor is used. This is the central-patch crop method. The second method, which is five-patch crop, is to crop five 224 by 224 image patches out of the image: from the four corners and the central area, and then the mean descriptor of these five image patches is used.

*4.4. Face Re-Identification*

In this paper, the Re-ID stage is treated as an online clustering problem: given a set of clustered data points (the gallery set G), the target is to find the matched clusters for the incoming data points (the probe set P), as shown in Fig. 2.

The gallery set G consists of multiple descriptors for each face, and is updated dynamically when new descriptors come into for Re-ID. Denote the gallery set as $G = \{G_1,G_2,G_3,\ldots,G_N\}$. N is the number of recorded IDs in gallery, and $G_x$ refers to the descriptor set of face x, i.e. $G_x = \{E_{x1},E_{x2},\ldots,E_{xe}\}$ where $E_{xi}$ represent the $i^{th}$ recorded descriptor of face x in gallery. The probe set is referred as $P = \{P_1,P_2,\ldots,P_M\}$ where M is the number of probes and $P_i$ is the descriptor of the $i^{th}$ face for Re-ID. In order to find the matched IDs, the Density-Based Spatial Clustering (DBScan) [26] algorithm is implemented in this stage.

Let $P_x$ stands for a certain probe descriptor, then a distance matrix D between $P_x$ and G will be computed: $D = \{D_{x1},D_{x2},D_{x3}\ldots,D_{xN}\}$. In D, $D_{xi}$ is a set of Euclidean Distances between $P_x$ and the descriptor set of the $i^{th}$ face in the gallery, and it can be denoted as $D_{xi} = \{Eu(P_x,E_{i1}), Eu(P_x,E_{i2}), \ldots, Eu(P_x,E_{ie})\}$ where $Eu(P_x,E_y)$ represent the Euclidean Distance between descriptor $P_x$ and $E_y$. If $Eu(P_x,E_y) \leq T_d$, then $P_x$ and $E_y$ could be considered as neighbours. Based on DBScan algorithm, if we denote the distance threshold as $T_d$ and the neighbour number threshold as $T_n$, in order to determine if $ID(P_x) \subseteq ID(G)$, Eq.(4) should be satisfied:

$$\max_{i=1,2,\ldots,N}(|S_i|) \geq T_n, S_i = \{d \leq T_d \mid d \in D_{xi}\} \qquad (4)$$

In Eq.(4), $S_i$ is a subset of $D_{xi}$ where the Euclidean distance is smaller than $T_d$, and $|S_i|$ is the number of $P_x$'s neighbours in $G_i$. $T_d$ refers to the distance threshold in DBScan, while $T_n$ is the neighbour number threshold in DBScan. Therefore, if Eq.(4) is satisfied, $P_x$ is close enough to a certain face in the gallery and it could be claimed that $ID(P_x) \subseteq ID(G)$. Assume $ID(P_x)=ID(G_{i^*})$, then $i^*$ could be determined using Eq.(5). Eq.(5) finds in the gallery a face which has the maximum number of neighbours with $P_x$, or in other words, the top ranked ID in the gallery. If two or more top ranked IDs are found, the one with the smallest average distance will be used.

$$i^* = \arg\max_{i=1,2,\ldots,N}(|\{d \leq T_d \mid d \in D_{xi}\}|) \qquad (5)$$

If Eq.(4) is not satisfied, it could be reasonably deduced that $ID(P_x) \nsubseteq ID(G)$， thus a new ID will be registered in the gallery and will be assigned to $P_x$.

After $P_x$'s assigned ID has been determined, $P_x$ will be removed from probe set and the same process will be repeated for the remaining probe descriptors. However, it should be noticed that Eq.(3) also should be satisfied, i.e. the given IDs for each probe element should be mutual exclusive. Therefore, assume $ID(P_x)=ID(G_y)$ , when processing $P_{x+1}$, the descriptor set $G_y$ will be ignored when computing the distance matrix D, so that $P_{x+1}$ cannot be assigned to the same ID. This will also help to speed up the Re-ID stage. When IDs for the probe set have been determined, the final stage is to return these IDs and to update the gallery using the probe set.

*4.5. Gallery Update and Management*

When the Re-ID system is initialised, the gallery set is empty, i.e. no supervised information is available. In this case, when new descriptors $\{P_1,P_2,\ldots, P_N\}$ arrives, they will be simply assigned with different IDs and will be added into the gallery. In all other cases, for a probe $P_x$, two different update processes are implemented:

1. If $ID(P_x) \subseteq ID(G)$, assume $ID(P_x)=ID(G_Y)$, then $P_x$ will be added to $G_Y$.
2. If $ID(P_x) \nsubseteq ID(G)$, then a new descriptor $G_{N+1}$ set will be created in gallery, and $P_x$ will be added to $G_{N+1}$.

In realistic scenarios, the gallery set will keep growing as more probe sets come into for Re-ID, while the size of the gallery set cannot be unlimited due to the real-time requirement. Therefore, the management of the gallery's size is necessary and essential. In this Re-ID system, the size of the gallery set is examined regularly to assure it has not overgrown. Particularly, for a gallery set $G = \{G_1,G_2,G_3,\ldots,G_N\}$, the following conditions should be checked:

$$\max_{i=1,2,\ldots,N}(|G_i|) \leq S_1 \;\&\; N \leq S_2 \qquad (6)$$

Eq.(6) poses restrictions on both the maximum number of descriptors ($S_1$) for $G_i$ and the maximum ID number ($S_2$) in the gallery. If any of these two restrictions are not true, corresponding reductions will be made on the gallery set to achieve Eq.(6), e.g. remove the number of $|G_k|-S_1$ descriptors out of $G_k$ if $|G_k|>S_1$.

**5. Evaluation Approaches**

Currently, the most widely-applied evaluation metric for person Re-ID problem is the called the Cumulative Matching Characteristic (CMC) curve. Traditionally the Re-ID problem is treated as a ranking problem: the candidates in the gallery are ranked by their similarities to the probe, and CMC (k) evaluates the probability that the true matching could be found in the top-k ranked candidates in the gallery. Although CMC is an applicable choice for close-set Re-ID problems, it does not suit the open-set Re-ID scenarios where the novelty detection should be made first.

So far the research on open-set Re-ID problems is limited compared to the close-set ones, and there are still several open evaluation issues. Different from close-set issues, the probe's IDs are not essentially a subset of the gallery, therefore the performance of novelty detection should also be evaluated. [27][28] propose an open-set Re-ID evaluation approach which measures both the Re-ID accuracy and the False Acceptance Rate (FAR). The Re-ID accuracy refers to how well the system can find the correct ID for a given probe in the gallery (true positive).

The FAR is expressed by two factors: the mismatches (MM) which describe the incorrect gallery matchings when the probe's ID is in the gallery, and False Positives (FP) where the probe is matched to the gallery while it is not part of it. The performance is measured by the curve of Re-ID accuracy vs. FAR rate.

This approach, however, is also not an appropriate option for our proposed scenarios. Different from [27][28] where a gallery set with relatively large size is provided in advanced and is not updated during the Re-ID process, our system is initialised with an empty gallery set which is updated gradually and dynamically using the probe set. Therefore, a False Positive probe could degrade the purity of the gallery set by the registration of a False Positive ID, and one tricky problem will emerge: the matched IDs found in the gallery are no longer 'trusty' ones. For example, one face which has already been recorded as ID 2 in the gallery is mistakenly recognised as a new ID 3. In this case, ID 3 will be registered in the gallery. Later, when this face comes into for Re-ID again, it could be potentially recognised as ID 2 or ID 3. So determining which ID is correct and how to measure the performance of this system is an open problem.

In this work, we propose a new evaluation approach to suit our scenarios: the Consistency Confusion Matrix (CCM) which measures the consistency of the predicted results. Specifically, in CCM, no matter how many IDs are assigned to a face, the major one will be recognised as the 'correct' one. This approach is actually evaluating the Re-ID system's abilities of consistently re-identifying the same face as the same ID without reference to the gallery set.

**Table 1**. An example of a 3 by 5 CCM (excluding the titles). Each row corresponds to an actual face, and each column represents a predicted ID. The 'unknown' column refers to faces that are not successfully recognised.

| | Predicted ID 4 | Predicted ID 1 | Predicted ID 3 | Predicted ID 2 | Unknown |
|---|---|---|---|---|---|
| Actual Face A | 32 | 0 | 0 | 27 | 3 |
| Actual Face B | 4 | 196 | 0 | 0 | 3 |
| Actual Face C | 0 | 0 | 128 | 0 | 4 |

Table 1 demonstrates an example of the CCM, which is a 3 by 5 matrix. Each row of the matrix is the actual face, while each column except the last one represents the predicted ID by the Re-ID system. For example, face A is predicted to be ID 2 and 4, while the major ID 4 is recognised as the matched ID for face A. The last column, which is named 'unknown', represents the faces that are not successfully recognised by this system due to the nature of DBScan algorithm and other potential reasons like the missing of frames.

In CCM, the sequence of the column is sorted so that the True Positives (TP) can be found in cells {i, i}. For instance, for the $i^{th}$ row (the actual face) of a P by Q CCM, the $i^{th}$ cell contains the True Positive (TP), while the $1^{st}$ to $P^{th}$ cells excluding the $i^{th}$ one represent the False Acceptance (FA) where the current face is assigned a major ID of other faces. The $(P+1)^{th}$ to the $Q^{th}$ cells correspond to the False Rejection (FR) where the probe is given a new ID although its correct ID already exist in the gallery. The overall TPs can be found by Eq.(7), and the Re-ID accuracy of CCM could be calculated as: Accuracy = TP/(sum(CMC)), where sum(CMC) refers to the sum of all CMC's elements.

$$TP=\sum_{i=1}^{P} CMC(i,i) \quad (7)$$

In this paper, the Re-ID accuracy is the main evaluation approach for the Re-ID system's performance. In addition to the Re-ID accuracy, False Acceptance Rate (FAR) and False Rejection Rate (FRR) are also important evaluation protocols and can be obtained from the CCM. As mentioned before, for each actual face, we have defined its major predicted ID as the correct one, therefore for a P by Q CCM, we can calculate the number of False Acceptance (FA) and False Rejection (FR) examples as in Eq.(8) and (9). Then the FAR and FRR can be obtained by: FAR = FA/(sum(CMC)) and FRR = FR/(sum(CMC)).

$$\text{FA}=\sum_{i=1}^{P}\sum_{j=1}^{P} CMC(i,j) \ , j \neq i \quad (8)$$

$$\text{FR}=\sum_{i=1}^{P}\sum_{j=P+1}^{Q} CMC(i,j) \quad (9)$$

As discussed in [16][19], FAR can affect the performance of a re-identification system more significantly than FRR. Classifying multiple persons' faces into one cluster may cause the robot to consistently re-identify their following faces into this incorrect class, while it generally harms less if the same person's face is categorised into different classes. Therefore, despite the trade-off between these two parameters, to achieve a lower FAR is practically more important than a lower FRR for this Re-ID system.

## 6. Experiments

Two sets of experiments are described in this section.

The target of the first experiment is to test different pre-processing approaches and evaluate the performance in a challenging dataset collected in the wild. This experiment is operated on a video dataset called TERESA dataset, which is collected by our TERESA robot. Another set of experiments is conducted on the YouTube Faces (YTF) dataset. The purpose is to test our Re-ID system on a widely-used face video dataset.

### *6.1. Experiments on TERESA Dataset*

*a. Dataset*

**Table 2:** Overview of the TERESA video dataset

| Index | Total No. of Faces | Duration |
|---|---|---|
| C1 | 2 | 107 Sec. |
| C2 | 4 | 37 Sec. |
| C3 | 3 | 58 Sec. |
| C4 | 3 | 138 Sec. |
| C5 | 3 | 110 Sec. |
| C6 | 7 | 153 Sec |
| C7 | 4 | 186 Sec |

TERESA is an intelligent social robot built to benefit the everyday life of elder people. We have created a database [29] which records the daily interactions between the TERESA robot and its users. We have manually selected seven videos out of the database and we have annotated each frame in terms of face identities. The total duration time is 13 Minutes and 12 Seconds, and Table 2 shows the overall information for each video clip.

Video clip C1 is captured by a static camera recording the robot's controller, while clips C2~C7 have been recorded by the robot camera while the robot interacts with elder people. The robotic videos are more challenging than the camera video due to the frequent movement of robot. As shown in Fig. 4, each video clip demonstrates a representative scenario which our robot may encounter with. For instance, C2 and C3 includes scenes where the users appear in front of the robot one after the other, while in C7 there are scenes where multiple faces appear simultaneously most of the time.

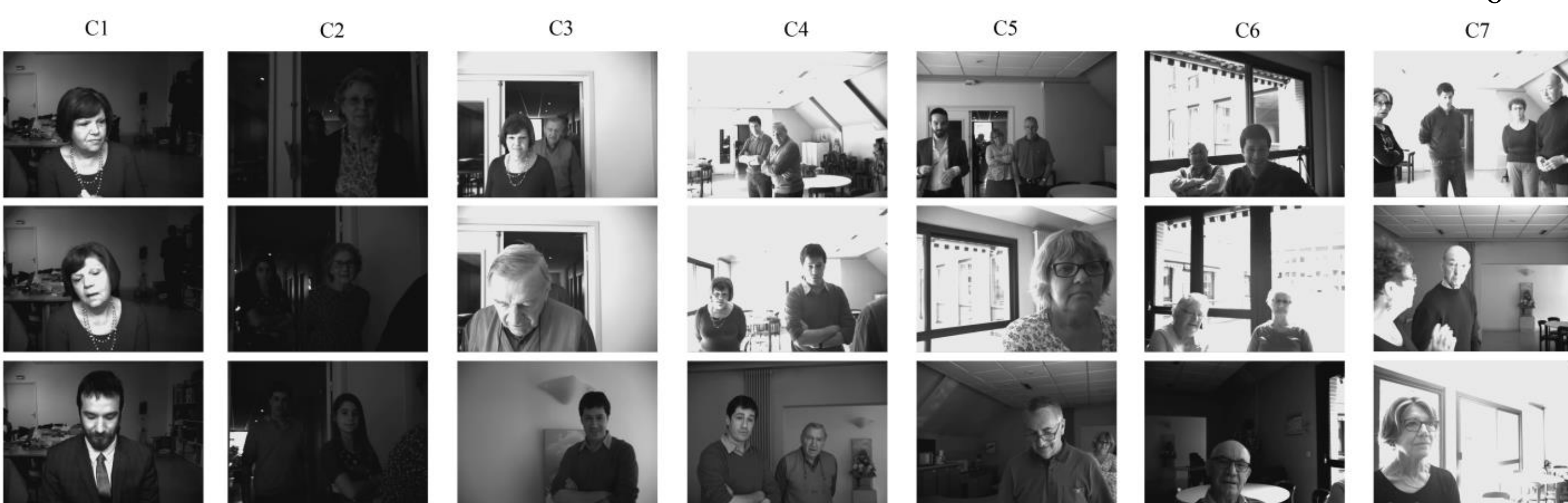

**Fig. 4.** Some scenarios in each video clip

All seven videos clips are taken in an unconstrained in-door environment during the interaction between TERESA and its users, and there are several challenging scenarios where the frame images are blurry and the identities of faces are varying fast due to the rapid movement of TERESA robot. Additionally, there are occasions where multiple faces are present simultaneously. What's more, the lighting conditions are also not consistent, e.g. C2 is much darker than other clips.

The resolution of all video clips which are black-and-white is uniformly reduced to 700 by 512 and the frame rate is 32 Frames Per Second (FPS).

*b. Experimental Setup*

Specifically, we have used the video clips C2 and C7 as the validation set to fine tune the hyper-parameters, and tested the performance of the Re-ID system on the rest five video clips. In order to optimise the Re-ID system and to improve its performance, we have tested various techniques at different stages of the pipeline, which are listed in Table 3.

**Table 3:** An overview of the tested techniques

| Pipeline Stage | Tested Techniques |
|---|---|
| Pre-Processing | Ghost Elimination, Histogram Equalization, Face Alignment |
| Feature Extraction | Central-patch / Five-patch Cropping |
| Re-ID | Online Clustering: DBScan, k-Nearest-Neighbours;<br>Offline Clustering: Gaussian Mixture Model (GMM), Hierarchical Clustering, K-Means Clustering, Spectral Clustering |
| Gallery Scalability | Size Limitation / No Size Limitation |

As can be seen from Table 3, for the pre-processing step, we have tested how the application of ghost elimination and histogram equalization can influence the Re-ID accuracy.

Additionally, we also evaluate if the application of face alignment enhances the performance. The alignment technology provided by [20] is used. Particularly, if the frontal angle of the face before alignment is larger than 15 degree, aligned faces with soft symmetry will be used, otherwise the alignment with no symmetry will be applied.

The feature extraction step is implemented with the Matlab toolbox MatConvNet [30] where the GPU mode is enabled to accelerate the processing speed. Also the central-patch cropping and five-patch cropping approaches are evaluated.

In the Re-ID step, there are two kinds of clustering problems depending on the environmental settings: the offline clustering problem where all the data points to be clustered are available at the same time, and the online clustering problem in which the data points are provided by batches.

The proposed scenarios distinctly fall into the online clustering problem. However, for simplicity, various clustering algorithms such as GMM, K-Means, DBScan are measured under offline settings in order to determine the most suitable algorithm. Therefore, we first implemented an offline clustering experiment where various offline clustering algorithms in Table 3 are evaluated.

Specifically, for this offline clustering algorithm test, all the available faces (assume F faces) in a video clip are extracted using the five-patch crop approach, and F descriptors will be obtained where each descriptor could be seen as a data point with 4096 dimensions. Then the offline clustering algorithms in Table 3 are applied to cluster these data points. For algorithms which require the cluster number as the parameters, the cluster number (which is the number of facial identities) is given. For GMM and DBScan, their hyper-parameters are fine-tuned with another video clips (descriptors are also extracted using five-patch crop), and their performance are evaluated on the same video clip like other algorithms.

The most appropriate algorithm is determined after the offline clustering algorithm test and is converted into online version. Additionally another online clustering algorithm, the k-Nearest-Neighbours where k=1, is also evaluated. Particularly, the CCMs of video clip C1, C3~C6 are obtained separately, i.e. the gallery set is empty at the start of each clip, and the overall accuracy is calculated by the sum of all five CCMs' TPs divided by the sum of all five CCMs' elements. The overall FAR and FRR are calculated using the same method.

For the gallery scalability, the parameters $S_1$ and $S_2$ in Eq.(6) are set to 60 and 20 respectively.

All our experiments are implemented on a computer with GTX 970M (3G) Graphic Card, i7-6700HQ (2.6GHz) CPU and 16G memory.

*c. Experimental Results*

First the result about the offline clustering algorithm will be reported. As shown in Table 4, Hierarchical Clustering and DBScan are the two top-performing ones. For Hierarchical Clustering to work, the cluster number should be determined in advance. However our proposed scenarios are strictly unsupervised and it will be difficult to gather user information in advance. For DBScan, what needs to be determined beforehand is the neighbourhood size and the distance threshold, which could be obtained by the validation process. Therefore, DBScan is more suitable for our proposed scenarios, and is employed in the Re-ID system.

**Table 4:** The performances of various offline clustering algorithms

| Clustering Algorithm | Required Parameters | Accuracy (%) |
|---|---|---|
| Hierarchical Clustering | Cluster Number | 99.06 |
| DBScan | Distance Threshold, Neighbourhood Size | 98.34 |
| k-Means | Cluster Number | 84.43 |
| Gaussian Mixture Model | Various | 79.59 |
| Spectral Clustering | Cluster Number | 62.87 |

Table 5 demonstrates the achieved accuracy, FAR and FRR of various techniques for this Re-ID system. As can be seen from the table, if no ghost elimination is applied, only 71.37% accuracy is achieve with a FAR as high as 21.45%. The application of Ghost Elimination increases the accuracy to 92.67% and significantly reduces the FAR. It can also be seen that the DBScan algorithm outperforms k-Nearest-Neighbours (k=1) which results in accuracy of 83.24% and a 16.00% FAR. A possible reason is that the DBScan algorithm requires the presence of multiple neighbours for a probe to be assigned an ID, therefore it is more tolerant to outliers than 1-Nearest-Neighbours.

The five-patch approach slightly surpasses the performance of the central-patch, achieving an accuracy of 92.81% with acceptable FAR and FRR, but it is much more time-consuming. Considering the real-time requirement, we have opted for the central-patch approach.

The employment of gallery size limitation deduces the accuracy to a small extent compared with unlimited gallery size, which is acceptable for realistic applications. As for the application of Histogram Equalization, it slightly increases the accuracy by 0.88% and reduces the FRR by 0.89%. The face alignment slightly reduces the accuracy from 93.55% to 92.53%, leading to a higher FAR as well. This could be possibly attributed to the artifacts introduced in the facial images due to frontalisation. Considering the time requirement, the face alignment technique is not applied in the following experiment.

Therefore, the optimised combination of techniques for this Re-ID system is established as: Ghost Elimination with Histogram Equalization at Pre-process stage, Central-Patch for feature extraction, DBScan for Re-ID step and size limitation as gallery scalability strategy.

Additionally, the values of hyper-parameters, like $T_d$ and $T_n$ in Eq.(4), could significantly affect the performance, and the fine-tuning of such parameters is also critical. Fig. 5 shows the impact of the distance threshold ($T_d$ in Eq.(4)) on the validation set (Video Clip C2 and C7) performance. $T_d$ is one of the most important hyper-parameters, as it is the distance threshold for determining if two descriptors are neighbours. As can be seen, the value of $T_d$ can greatly influence the Re-ID accuracy, FAR and FRR. Despite the fluctuations resulted from the relatively small number of validation samples, it can be observed that the accuracy curve peaks when $T_d$ is between 1.19 and 1.225, a range where the FAR and FRR curves have also achieved a relatively balanced state. As discussed before, for a Re-ID system with empty initialising database, FAR is more important than FRR, therefore we have empirically set the $T_d$ value to 1.215, where an accuracy of around 92.4% and a comparatively low FAR (approximately one fourth of FRR) are achieved.

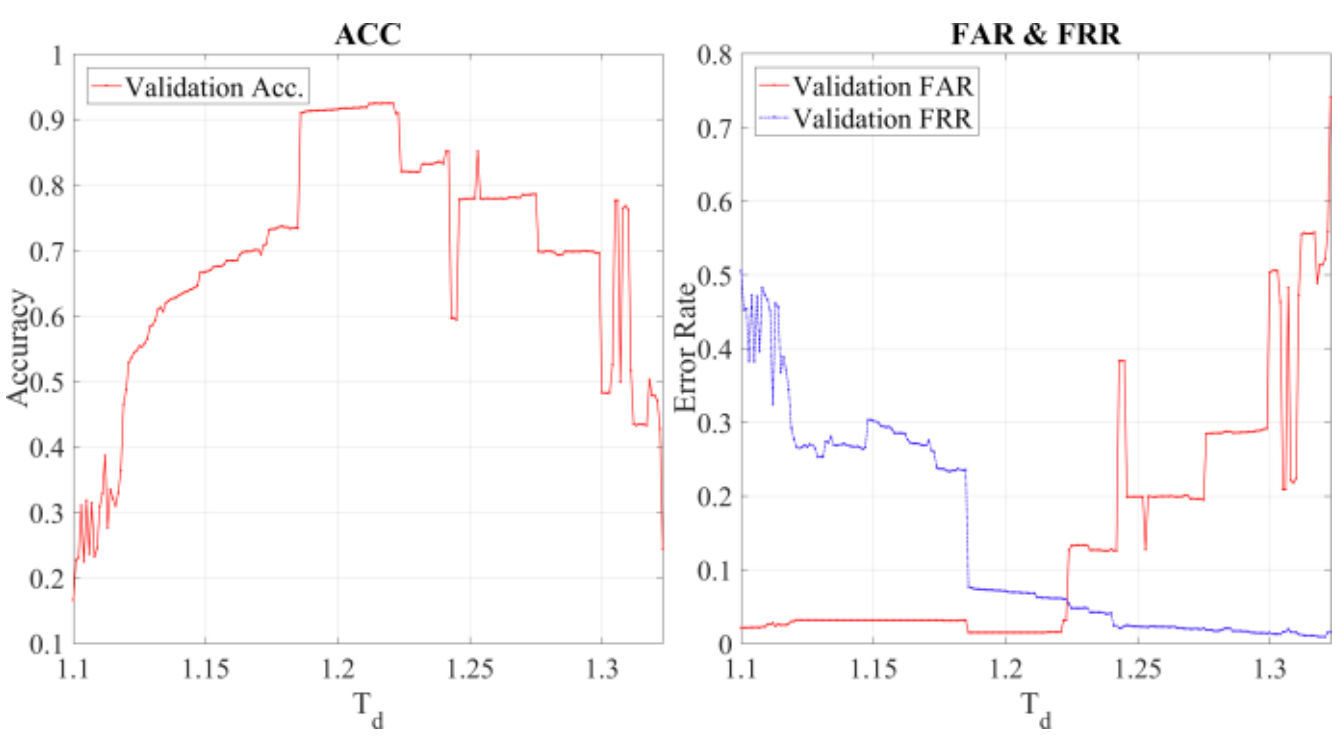

**Fig. 5.** Effect of distance threshold $T_d$ on accuracy (LEFT) and FAR & FRR (RIGHT) (Best seen in colour)

Fig. 6 plots the relationships between the accuracy on validation set and two other hyper-parameters: the neighbour number threshold $T_n$ in Eq.(4), and the limitation number $S_l$ in Eq.(6) for gallery scalability, which constrains the maximum number of descriptors per ID to keep in the gallery. We can see that the accuracy is maximised when $T_n$ is 3, after which it gradually decreases. A possible reason may be that a large $T_n$ requires more faces to be placed in the database before a valid ID could be assigned, leading to an increase of unrecognised faces. As a result, the value of $T_n$ is set to 3.

**Table 5:** The performances of various techniques

| Pre-Processing & Alignment | Feature Extraction | Re-ID | Gallery Scalability | Accuracy (%) | FAR (%) | FRR(%) |
|---|---|---|---|---|---|---|
| None | Central-patch | DBScan | Size Limitation | 71.37 | 21.45 | 7.18 |
| Ghost Elimination | Central-patch | DBScan | Size Limitation | 92.67 | 2.77 | 4.56 |
| Ghost Elimination | Central patch | k-Nearest-Neighbour (k=1) | Size Limitation | 83.24 | 16.00 | 0.76 |
| Ghost Elimination | Five-Patch | DBScan | Size Limitation | 92.81 | 2.73 | 4.46 |
| Ghost Elimination | Central patch | DBScan | Unlimited DB Size | 92.84 | 2.71 | 4.43 |
| Ghost Elimination & Histogram Equalization | Central-patch | DBScan | Size Limitation | 93.55 | 2.78 | 3.67 |
| Ghost Elimination & Histogram Equalization & Face Alignment | Central patch | DBScan | Size Limitation | 92.53 | 4.58 | 2.89 |

As for $S_1$, an obvious increasing trend of accuracy can be observed when $S_1$ grows from 1 to 15, and then the accuracy curve generally remains stable. This trend indicates that the performance can be generally improved by keeping more descriptors per ID in the gallery set (up to 15), or in other words, recording more information for each person. Considering that the size of validation set is relatively small, we have empirically set $S_1$ to 60 for a better robustness under real-time scenarios. Another hyper-parameter is $S_2$, which is the maximum number of IDs to keep in the database. This is a flexible hyper-parameter as its value should be based on the applied scenarios. Here we just use the value of 20.

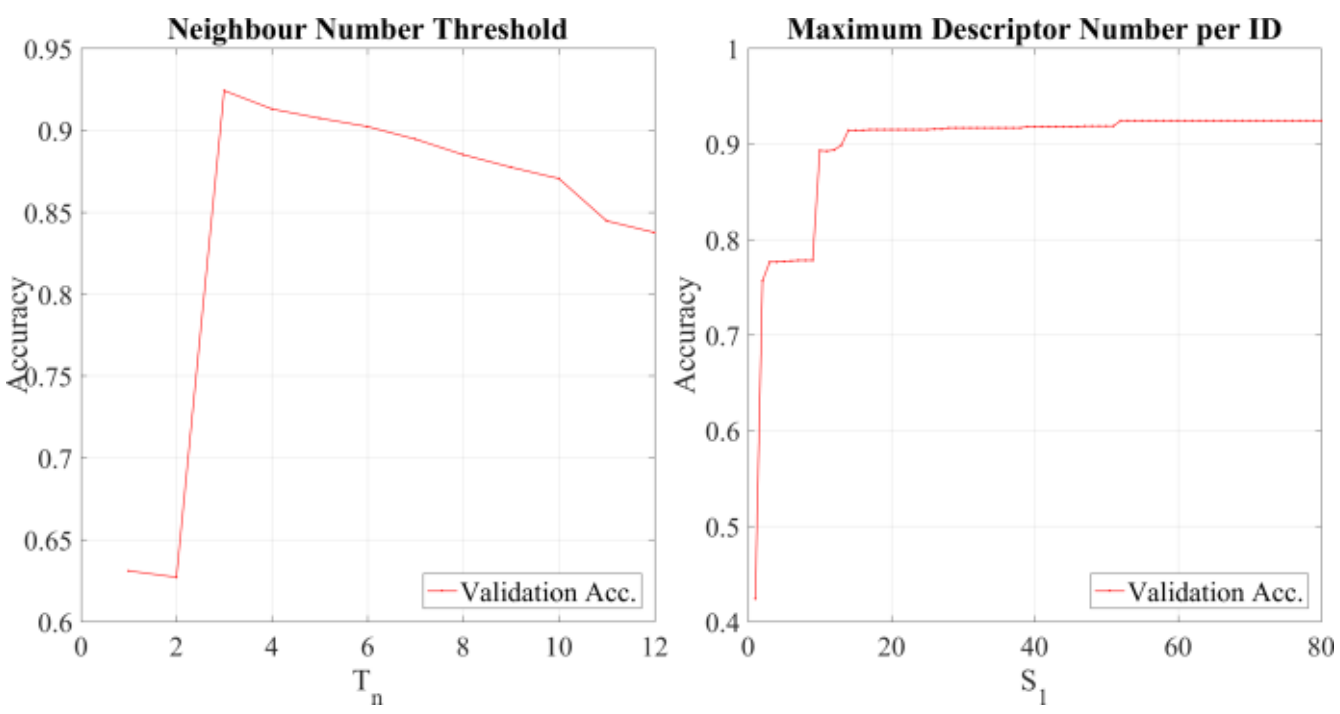

**Fig. 6.** Effect of neighbour number threshold $T_n$ and the limitation number $S_1$ on accuracy (LEFT) and FAR & FRR (RIGHT) (Best seen in colour)

Using the optimised technique combinations and those hyper-parameter values, the performances achieved for each video clip are demonstrated in Table 6.

**Table 6**: The achieved performances on each clip of TERESA dataset

| Clip Index | Total Faces No. | Accuracy (%) | FAR(%) | FRR(%) |
|---|---|---|---|---|
| C1 | 2 | 99.45 | 0.30 | 0.25 |
| C2 | 4 | 94.79 | 1.04 | 4.17 |
| C3 | 3 | 90.67 | 5.33 | 4.00 |
| C4 | 3 | 83.92 | 0.00 | 16.08 |
| C5 | 3 | 70.37 | 20.74 | 8.89 |
| C6 | 7 | 92.72 | 5.82 | 1.46 |
| C7 | 4 | 91.98 | 1.52 | 6.50 |
| Overall | | 93.55 | 2.78 | 3.67 |

Table 7: Processing time overview

| **Feature Extraction Approach** | Central-Patch | | Five-Patch | |
|---|---|---|---|---|
| **Gallery Size** | Empty | Full | Empty | Full |
| **Processing Time (Sec. Per Face)** | 0.038 | 0.095 | 0.17 | 0.25 |

Table 7 shows the processing time of different feature extraction approaches with the size limitation enabled (The maximum gallery size $S_1$ and $S_2$ in Eq.(6) are 60 and 20). The processing time refers to the time elapse between the input of a detected face image and the output of the predicted ID (The face detection time is not included), which is related to the feature extraction approaches and actual gallery size. It can be seen that our Re-ID system can achieve high processing speed, which satisfies the real-time requirement of HRI.

### *6.2. Experiments on YTF Dataset*

#### *a. Dataset*

The YouTube Face (YTF) [25] dataset is a widely-used benchmark for face recognition in the wild. There are 3,425 videos from 1,595 persons in this dataset, with an average of 2.15 videos per subject. The suggested evaluation protocol is the following: ten folds where each fold contains 250 matched video pairs and 250 mismatched video pairs. So there are a total number of 5,000 video pairs in the benchmark testing. The purpose of using this dataset is to test how this Re-ID system works on a public dataset of unconstrained videos with even lower resolution like 320 by 240 in YTF.

#### *b. Experimental Setup*

The target of this experiment is to test the optimised Re-ID system's performance on the YTF dataset. However, our proposed Re-ID system is not designed for face verification tasks, therefore the YTF's evaluation protocol cannot be applied directly and needs to be modified to suit our system.

Particularly, we have divided the 5,000 video pairs into 500 folds, each fold containing 5 matched and 5 mismatched video pairs, and the Re-ID system is operated on the 500 folds independently. The first 100 folds are selected as the validation set to fine tune the parameters of the Re-ID system, and the rest (fold 101 to 500) are the test set. As a result, $T_d$ and $T_n$ in Eq.(4) are set to 1.27 and 3 respectively. The overall accuracy and Unweighted Average Recall (UAR) of the test set are reported. We selected 10 video pairs in one fold in order to limit the number of subjects in each fold to around 15, and as a consequence keep the gallery size relatively small. This is a realistic assumption since the number of faces encountered by social robot will usually be limited to a threshold such as 15.

For each fold, the Re-ID will be initialized as an empty gallery set, and then the videos of this fold will be input into the Re-ID system one by one to obtain the CCM of this fold. The CCMs of the test set are obtained to calculate the overall accuracy and UAR.

#### *c. Experimental Results*

The average accuracy and UAR are shown in Table 8. In [7] an accuracy of 97.3% is reported on the YTF dataset unrestricted protocol, which is around 6.9% higher than ours. However, it should be noted that our accuracy is achieved for a more challenging task than unstrained face verification, which is to re-identify the faces from 10~20 people at one time rather than from 1~2 people. Considering the difficulty, the achieved accuracy and UAR are arguably good ones.

**Table 8**: Results on the YTF dataset

| Accuracy (%) | UAR (%) |
|---|---|
| 90.41 | 84.79 |

## 7. Software Implementation

We have implemented our proposed method into a software module in the HCI^2 Framework [1] in order for it to be further integrated into the TERESA robot [2]. Internality, the face Re-ID modules utilises GPU-enabled MatConvNet toolbox to compute the feature vectors, thus is able to achieve high processing speed in real world conditions. In particular, our tests show our implementation on-board the robot is able to perform the Re-ID task at a frame rate of 10~26 fps (depending on the gallery size) with an input video resolution of 700 by 512 pixels.

## 8. Discussion & Future Work

### *8.1. Performance*

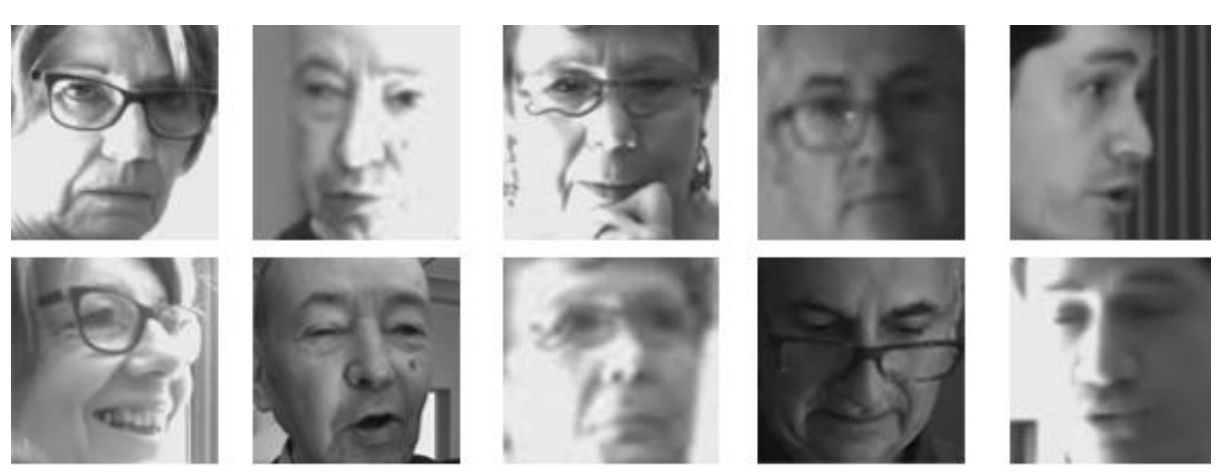

**Fig. 7.** Some false negative face pairs in TERESA dataset; each column are faces from the same person

Generally, it is observed that this Re-ID system is experiencing difficulties in re-identifying certain intra-class faces. In Fig. 7, several false negative face pairs on TERESA dataset are plotted. It can be shown that this Re-ID system tend to make incorrect re-identifications for intra-class faces if some faces' image qualities are comparatively low. Additionally, the varying expressions, poses or glasses could also increase the occurrences of incorrect re-identification.

Also it can be seen from Table 6 that for certain video clips in TERESA dataset such as C5, the achieved accuracy is comparatively lower, while for certain video clips like C1, the performance is very high. Fig. 8 shows some typical faces in C1 and C5, and it could be seen that the faces in C1 are generally clearer and easier to recognise than that of C5. C1 is recorded from a fixed camera while C5 is taken by the moving robot. In addition, the lighting environment also varies significantly in C5, while that of C1 remains stable.

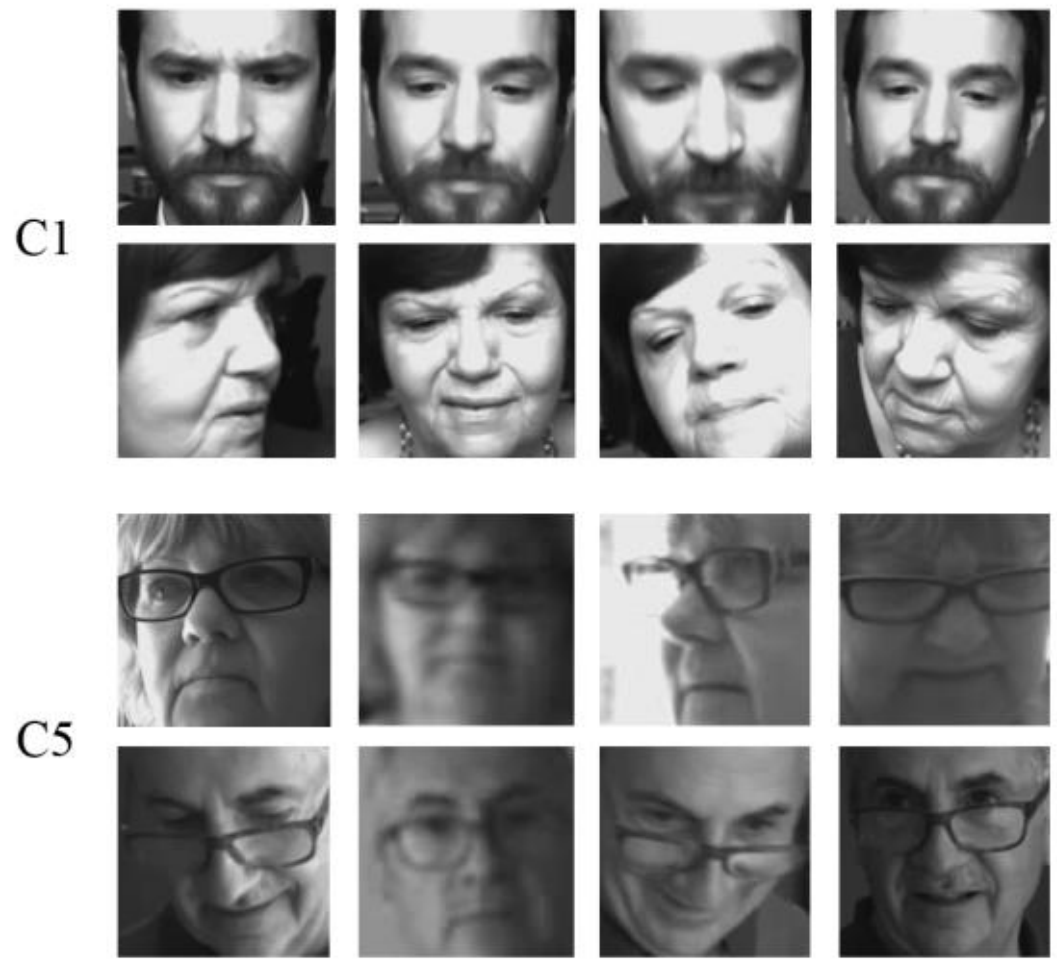

**Fig. 8.** Some face images from video clip C1 and C5. Each row is faces from the same subject. The first two rows are from C1, while the last two ones are from C5.

### *8.2. Efficiency*

It should be noticed that it is necessary to limit the size of the gallery set for realistic applications. However the limitation on gallery size affects the Re-ID systems' abilities to recognise various faces. For example, if the number of facial IDs in the gallery set has achieved the maximum value, then when a new face comes in, some previous IDs will be cleared from the gallery set. Therefore, the appropriate gallery management strategy should be determined depending on the applied scenarios.

Regarding the processing time, the two most time-consuming stages is the feature extraction and Re-ID (find the matched ID in the gallery). For each face, the time of feature extraction is fixed, while the time of Re-ID increases linearly with the size of the gallery. The general processing time is a O(N) performance. Therefore, it would be an interesting future topic to explore how to improve the efficiency of the Re-ID stage. So far each face is represented with multiple descriptors in the gallery, and this complicates the process of measuring similarities. For instance, it might be possible to compute only one or two representative descriptors for each face, or to find a face modelling approach where only certain parameters are needed to describe a face.

In addition, we also have considered the potential of utilising temporal information. Although the current Re-ID framework has achieved real-time running speed for videos in the wild, the temporal information of such videos is still not utilised, while the abilities of capturing such information could further reduce the processing time and increase its robustness against noise. Therefore, another future work is to enable the system to acquire temporal information from videos, and the Long Short Term Memory Networks (LSTMs) [31], which have already shown powerful capacities in utilising temporal information, may be a good choice.

## 9. Conclusion

In this paper, an effective, real-time and unsupervised face Re-ID system for HRI is presented. A new evaluation approach for open-set Re-ID problem with vacant initial gallery set is proposed. Also the optimised combination of techniques is reported through the experiments. Experimental results on TERESA dataset and YTF dataset demonstrate that this Re-ID system can achieve high accuracy and is capable of real-time execution. The analysis has shown that the varying image qualities, expressions, posed, etc. brought by the interaction between human and social robot increases the difficulties of face re-identification, and future work is required to further improve the performance and processing speed.

## 10. Acknowledgement

This work is funded by the EPSRC project EP/N007743/1 (FACER2VM). The work of Jie Shen and Stavros Petridis has also been partially supported by the European Community FP7 under grant agreement no. 611153 (TERESA).